\documentclass[
]{ceurart}

\usepackage{listings}
\usepackage{threeparttable}
\usepackage{tabularx}
\usepackage{color}
\usepackage{booktabs}
\usepackage{verbatim}
\usepackage{amsopn}
\usepackage{amsmath}
\usepackage{multirow}
\usepackage{colortbl}
\usepackage{makecell} 

\usepackage{eucal}
\usepackage{amsfonts}
\usepackage{bbm}
\usepackage{bm}

\usepackage{xspace}

\makeatletter
\DeclareRobustCommand\onedot{\futurelet\@let@token\@onedot}
\def\@onedot{\ifx\@let@token.\else.\null\fi\xspace}

\def\eg{\emph{e.g}\onedot} 
\def\ie{\emph{i.e}\onedot}

\makeatother

\begin{document}

\copyrightyear{2023}
\copyrightclause{Copyright for this paper by its authors.
  Use permitted under Creative Commons License Attribution 4.0
  International (CC BY 4.0).}

\conference{MiGA@IJCAI23: International IJCAI Workshop on Micro-gesture Analysis for Hidden Emotion Understanding, August 21, 2023, Macao, China.}

\title{Joint Skeletal and Semantic Embedding Loss for Micro-gesture Classification}

\author[1]{Kun Li}[%
orcid=0000-0001-5083-2145,
email=kunli.hfut@gmail.com,
]

\author[1,2,3]{Dan Guo}[%
orcid=0000-0003-2594-254X,
email=guodan@hfut.edu.cn,
]
\cormark[1]

\author[1]{Guoliang Chen}[%
orcid=0009-0002-7984-3184,
email=guoliangchen.hfut@gmail.com,
]

\author[1]{Xinge Peng}[%
orcid=0009-0006-3820-8810,
email=xg.pengv@gmail.com,
]

\author[1,2,3]{Meng Wang}[%
orcid=0000-0002-3094-7735,
email=eric.mengwang@gmail.com,
]
\cormark[1]

\address[1]{School of Computer Science and Information Engineering, School of Artificial Intelligence, Hefei University of Technology (HFUT)}
\address[2]{Key Laboratory of Knowledge Engineering with Big Data (HFUT), Ministry of Education}
\address[3]{Institute of Artificial Intelligence, Hefei Comprehensive National Science Center, China}

\cortext[1]{Corresponding authors.}

\begin{abstract}
In this paper, we briefly introduce the solution of our team HFUT-VUT for the Micros-gesture Classification in the MiGA challenge at IJCAI 2023. 
The micro-gesture classification task aims at recognizing the action category of a given video based on the skeleton data. 
For this task, we propose a 3D-CNNs-based micro-gesture recognition network, which incorporates a skeletal and semantic embedding loss to improve action classification performance.  
Finally, we rank \textbf{1st in the Micro-gesture Classification Challenge, surpassing the second-place team in terms of Top-1 accuracy by 1.10\%}. 
\end{abstract}


\maketitle

\section{Introduction}
Micro-gesture Analysis for Hidden Emotion Understanding (MiGA) is a Challenge at IJCAI 2023. It is launched based on the iMiGUE~\cite{liu2021imigue} and SMG~\cite{chen2023smg} datasets and requires understanding emotion based on the micro-gestures (MGs). 
The micro-gesture classification challenge aims to recognition MGs from short video clips based on the skeleton data. The iMiGUE dataset were collected from post-match press conferences. Compared to ordinary action or gesture recognition, MGs present more challenge. MGs encompass more refined and subtle bodily movements that occur spontaneously during real-life interactions. Additionally, there is an imbalanced distribution of MGs, where 28 out of 32 categories accounted for 57.8\% of the data. 
In this challenge, we adopt the skeleton-based recognition model PoseC3D~\cite{duan2022revisiting} as the baseline model, and introduce semantic embedding of action label~\cite{frome2013devise,yeh2019multilabel,wei2020learning,filntisis2020emotion} to supervise the action classification. 

The main contributions of our method are summarized as follows. 
\begin{itemize}
\item We proposed a CNN-based network for micro-gesture classification. Specifically, we incorporate skeletal and semantic embedding loss for action classification. 
\item For the micro-gesture classification challenge, our method achieves a Top-1 accuracy of 64.12 on the iMiGUE test set. 
For the SMG dataset, our proposed method achieves 68.03 and 94.76 of Top-1 and Top-5 accuracy, respectively. 
The experimental results indicate that our method effectively captures subtle changes of micro-gestures. 
\end{itemize}

\begin{figure}[t]
\centering
\includegraphics[width=1.0\linewidth]{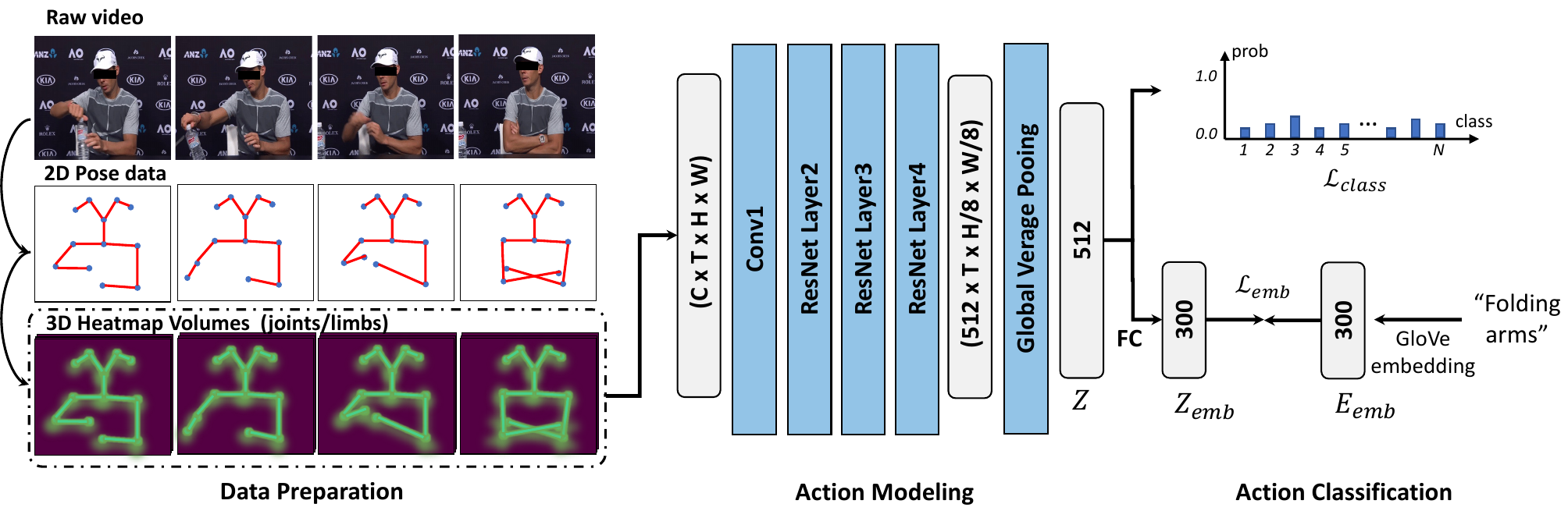}
\caption{Overview of the proposed method for micro-gesture classification. The proposed method consists of three key steps: data preparation, action modeling, and action classification.}
\label{fig:main}
\end{figure}

\section{Methodology}
\subsection{Data preparation}
Considering that there are no lower body actions in the iMiGUE dataset, only the 22 key points extracted by OpenPose toolbox~\cite{8765346} of the upper body are used. 
For the SMG dataset, 25 keypoints of whole body are used. 
As shown in Figure~\ref{fig:main},  given a video $\textbf{V}$, the extracted 2D pose data is denoted as $\textbf{X}\in\mathbb{R}^{T\times K\times C}$, where $T$ denotes the total frames, and $K$ denotes the number of keypoints, and $C$ is the number of dimension for keypoint coordinates. 
Then, we transform the 2D pose data $\textbf{X}$ to a 3D
heatmap volume with the size of $C\times T\times H \times W$. $C$ is the number of joints, $H$ and $W$ are the height and width of the heatmap. 
Finally, the subjects-centered cropping and uniform sampling strategies are used to reduce the redundancy of 3D heatmap volumes. More details about the 3D heatmap volumes can refer to PoseC3D~\cite{duan2022revisiting}. 

\subsection{Action modeling and classification} 
After getting the 3D heatmap volumes, here, we use 3D-CNNs to capture the spatiotemporal dynamics of skeleton sequences. Specifically, we first use the SlowOnly~\cite{feichtenhofer2019slowfast} model as the backbone for skeleton-based action recognition. Then, we use global average pooling to generate a skeletal embedding $Z\in \mathbb{R}^{512}$. 
The vector $Z$ is fed into a fully-connected (FC) layer for action classification. 
We also consider GloVe embedding of action label for the supervision of action classification. 
Specifically, we first transform the action label to 300-dimension GloVE~\cite{pennington2014glove} word embedding $E_{emb}$. Then, we use a fully-connected layer to convert the vector $Z\in \mathbb{R}^{512}$ to a 300-dimension vector $Z_{emb}\in\mathbb{R}^{300}$. Here, we use a semantic loss $\mathcal{L}_{emb}$ to make $Z_{emb}$ close to semantic embedding $E_{emb}$.

\subsection{Loss Optimization}
\begin{equation}
\mathcal{L}= \mathcal{L}_{class} + \alpha\cdot\mathcal{L}_{emb},
\label{eq:loss}
\end{equation}
\begin{equation}
\mathcal{L}_{emb} = ||Z_ {emb} - E_{emb}||^2,
\end{equation}
where $\alpha$ is a hyper-parameter to balance the two losses, and we will discuss it in the experiment. $\mathcal{L}_{emb}$ is MSE loss to supervise the semantic embedding. $\mathcal{L}_{class}=\mathcal{L}_{XE}$ is the cross-entropy loss to supervise the skeletal embedding. In addition, $\mathcal{L}_{class}$ also serve as the classification loss. 

\section{Experiments}
\subsection{Datasets}
\textbf{iMiGUE~\cite{liu2021imigue} dataset.} This dataset comprises 32 MGs, along with one non-MG class, collected from post-match press conferences videos of tennis players. 
This challenge follows a cross-subject evaluation protocol, wherein the 72 subjects are divided into a training set consisting of 37 subjects and a testing set comprising 35 subjects. 
For the MG classification track, 12,893, 777, and 4,562 MG clips from iMiGUE are used for train, val, and test, respectively. 
\textbf{SMG~\cite{chen2023smg} dataset.} This dataset consists of 3,692 samples of 17 MGs. The MG clips are annotated from 40 long video sequences, which in total contain 821,056 frames. Each long video sequence has a duration of 10-15 minutes. 
The dataset was collected from 40 subjects while narrating both a fake and a real story to elicit various emotional states. 

\subsection{Evaluation Metrics and Implementation Details}
For the micro-gesture classification challenge, we calculate the Top-1 Accuracy to assess the prediction results. 
For the micro-gesture classification challenge, we implement our approach with the PYSKL toolbox~\cite{duan2022pyskl}. The model is trained with SGD with momentum of 0.9, weight decay of 3e$^{-4}$. We set the batch size to 32, set the initial learning rate to 0.2/3. In addition, the model is trained 100 epochs with CosineAnnealing learning rate scheduler. 
The SlowOnly model is adopted as the 3D-CNN backbone. 
For the ensemble model (Joint\&Limb), we use the weighted summation of scores for two modalities with a ratio of 2:3.

\subsection{Experimental Results}
As shown in Table~\ref{tab:codalab_results}, we report top-3 results on the test set of the iMiGUE dataset. Our team achieves the best Top-1 Accuracy of 64.12, which is higher than the runner-up by 1.10\%. 
In addition, we also compare our approach with different skeleton-based action recognition methods on the iMiGUE and SMG datasets. 
At first, we investigate the impact of hyper-parameter $\alpha$ in Eq.~\ref{eq:loss}. As shown in Table~\ref{tab:abl_alpha}, the proposed method achieves the best Top-1 accuracy when $\alpha=20$. Thus, we set $\alpha=20$ as the optimal setting in the following experiments. 
Secondly, as shown in Table~\ref{tab:main}, on the iMiGUE dataset, our method achieves the best Top-1 and Top-5 accuracy of 64.12 and 91.10, respectively. Compared with the baseline model PoseC3D, our method exhibits 2.74\% improvements in Top-1 accuracy with the joint feature as input. 
On the SMG dataset, our method also achieves the best performance (\ie, 68.03 and 94.76 of Top-1 and Top-5 accuracy). Compared with the PoseC3D model, our method achieves 2.63\% improvements in Top-1 accuracy in terms of joint feature. In addition, we can see that the ensemble model (Joint\&Limb) also shows significant performance improvement (\ie, 1.96\% and 2.96\% improvements on Top-1 
 and Top-5 accuracy compared with `Joint'). 

\begin{table}
\footnotesize
\caption{The top-3 results of micro-gesture classification on the iMiGUE test set. Data is provided by the Codalab competition page\protect\footnotemark[1]. 
}
\begin{tabular}{c|cc}
\toprule
Rank & Team & Top-1 Accuracy (\%) \\ \hline
1 & \textbf{Ours} & \textbf{64.12}  \\ \hline
2 & NPU-Stanford  & 63.02 \\
3 & ChenxiCui & 62.63 \\
\bottomrule
\end{tabular}
\label{tab:codalab_results}
\end{table}
\footnotetext[1]{The Codalab competition page: \href{https://codalab.lisn.upsaclay.fr/competitions/11758\#results}{link} }

\begin{table}
\caption{Ablation study results of $\alpha$ on the iMiGUE test set with joint features.}
\begin{tabular}{c|cc}
\toprule
Parameter    & Top-1 (\%) & Top-5 (\%)  \\ \hline
$\alpha$=1   & 59.58 & 90.05 \\
$\alpha$=10  & 60.37 & 90.03\\
$\alpha$=20  & \textbf{62.28} & \textbf{90.62} \\
$\alpha$=30  & 61.03 & 89.89 \\
$\alpha$=40  & 60.21 & 90.11   \\
$\alpha$=50  & 61.60 & 90.31 \\ \bottomrule
\end{tabular}
\label{tab:abl_alpha}
\end{table}

\begin{table}[]
\caption{The results of micro-gesture classification on the iMiGUE and the SMG test sets.} 
\begin{tabular}{c|c|cc|cc}
\toprule
\multirow{2}{*}{Method} & \multirow{2}{*}{Modality} &  \multicolumn{2}{c|}{iMiGUE dataset} &  \multicolumn{2}{c}{SMG dataset} \\
 & & Top-1 (\%) & Top-5 (\%)   & Top-1 (\%) & Top-5 (\%) \\ \hline
ST-GCN~\cite{yan2018spatial}   & Joint & 46.38 & 85.47 & 58.03 & 93.61 \\
ST-GCN++~\cite{duan2022pyskl}  & Joint & 49.56 & 85.09 & 58.03 & 93.61 \\
StrongAug~\cite{duan2022pyskl} & Joint & 53.13 & 87.00 & 62.79 & 92.62 \\
AAGCN~\cite{shi2020skeleton}   & Joint & 54.73 & 84.59 & 60.49 & 91.64 \\
CTR-GCN~\cite{chen2021channel} & Joint & 53.02 & 86.19 & 60.98 & 90.82 \\
DG-STGCN~\cite{duan2022dg}     & Joint & 49.56 & 85.09 & 65.57 & 90.82 \\ 
PoseC3D~\cite{duan2022revisiting} & Joint & 59.54 & 89.59 &63.44 & 88.20 \\
PoseC3D~\cite{duan2022revisiting} & Limb & 60.74 & 90.51 & 63.11 & 93.77  \\
\hline
Ours & Joint & 62.28 & 90.62 & 66.07 & 91.80   \\
Ours & Limb & 63.48 & 91.01 & 65.57 & 92.62   \\
\textbf{Ours} & Joint\&Limb & \textbf{64.12} & \textbf{91.10} & \textbf{68.03} & \textbf{94.76} \\
\bottomrule
\end{tabular}
\label{tab:main}
\end{table}

\section{Conclusions}
In this paper, we present our solution developed for the MiGA challenge hosted at IJCAI 2023. 
Our approach adopts the PoseC3D model as a baseline, incorporating both skeletal embedding loss and semantic embedding loss. 
By leveraging the joint and limb modality data, our approach achieved the first place with the top-1 and top-5 accuracy of 64.12 and 91.10, respectively. 
In the future, we plan to address the issues in this challenge from other perspectives, \eg, more robust network for human pose estimation, data augmentation for imbalanced data learning, RGB-based visual feature for micro-gesture recognition, and temporal context modeling~\cite{guo2019dense,li2021proposal} for capturing subtle changes of MG. 

\begin{acknowledgments}
This work was supported by the National Natural Science Foundation of China (72188101, 62020106007, and 62272144, and U20A20183), and the Major Project of Anhui Province (202203a05020011).
\end{acknowledgments}

\bibliography{sample-ceur}

\appendix

\end{document}